\pdfoutput=1

\documentclass[11pt]{article}

\usepackage[final]{acl}

\usepackage{times}
\usepackage{latexsym}

\usepackage[T1]{fontenc}

\usepackage[utf8]{inputenc}

\usepackage{microtype}

\usepackage{inconsolata}

\usepackage{graphicx}

\usepackage{amsmath}
\usepackage{amssymb}
\usepackage{multirow}
\usepackage{multicol}
\usepackage{graphicx}
\usepackage{enumitem}
\usepackage{hyperref}
\usepackage{tikz}
\newcommand*{\circled}[1]{\lower.7ex\hbox{\tikz\draw (0pt, 0pt)%
    circle (.5em) node {\makebox[1em][c]{\scriptsize #1}};}}

\title{Scaling Model and Data for Multilingual Machine Translation with Open Large Language Models}

\author{Yuzhe Shang\textsuperscript{1}{\thanks{~Equal contribution. Correspondence to: Pengzhi Gao <gaopengzhi@xiaomi.com>.}\thanks{This work was done while Yuzhe Shang was interning at the MiLM Plus team.}}, Pengzhi Gao\textsuperscript{2}{\footnotemark[1]}, Wei Liu\textsuperscript{2}, Jian Luan\textsuperscript{2}, Jinsong Su\textsuperscript{1} \\
 \textsuperscript{1}School of Informatics, Xiamen University, China \\
 \textsuperscript{2}MiLM Plus, Xiaomi Inc., Beijing, China, \\
\texttt{shangyuzhe@stu.xmu.edu.cn, jssu@xmu.edu.cn} \\
\texttt{\{gaopengzhi,liuwei40,luanjian\}@xiaomi.com}
}

\begin{document}

\maketitle

\begin{abstract}

Open large language models (LLMs) have demonstrated improving multilingual capabilities in recent years. In this paper, we present a study of open LLMs for multilingual machine translation (MT) across a range of languages, and investigate the effects of model scaling and data scaling when adapting open LLMs to multilingual MT through continual pretraining and instruction finetuning. Based on the Gemma3 model family, we develop MiLMMT-46, which achieves top-tier multilingual translation performance across $46$ languages\footnote{Arabic (\texttt{ar}), Azerbaijani (\texttt{az}), Bulgarian (\texttt{bg}), Bengali (\texttt{bn}), Catalan (\texttt{ca}), Czech (\texttt{cs}), Danish (\texttt{da}), German (\texttt{de}), Greek (\texttt{el}), English (\texttt{en}), Spanish (\texttt{es}), Persian (\texttt{fa}), Finnish (\texttt{fi}), French (\texttt{fr}), Hebrew (\texttt{he}), Hindi (\texttt{hi}), Croatian (\texttt{hr}), Hungarian (\texttt{hu}), Indonesian (\texttt{id}), Italian (\texttt{it}), Japanese (\texttt{ja}), Kazakh (\texttt{kk}), Khmer (\texttt{km}), Korean (\texttt{ko}), Lao (\texttt{lo}), Malay (\texttt{ms}), Burmese (\texttt{my}), Norwegian (\texttt{nb}), Dutch (\texttt{nl}), Polish (\texttt{pl}), Portuguese (\texttt{pt}), Romanian (\texttt{ro}), Russian (\texttt{ru}), Slovak (\texttt{sk}), Slovenian (\texttt{sl}), Swedish (\texttt{sv}), Tamil (\texttt{ta}), Thai (\texttt{th}), Tagalog (\texttt{tl}), Turkish (\texttt{tr}), Urdu (\texttt{ur}), Uzbek (\texttt{uz}), Vietnamese (\texttt{vi}), Cantonese (\texttt{yue}), Chinese (Simplified) (\texttt{zhs}), and Chinese (Traditional) (\texttt{zht}).}. Extensive experiments show that MiLMMT-46 consistently outperforms recent state-of-the-art (SOTA) models, including Seed-X \cite{cheng2025seedxbuildingstrongmultilingual}, HY-MT-1.5 \cite{zheng2025hymt15technicalreport} and TranslateGemma \cite{finkelstein2026translategemmatechnicalreport}, and achieves competitive performance with strong proprietary systems such as Google Translate and Gemini 3 Pro.\footnote{Models are released at \href{https://huggingface.co/collections/xiaomi-research/milmmt-46}{https://huggingface/MiLMMT}. Codes are released at \href{https://github.com/xiaomi-research/gemmax}{https://github/MiLMMT}.}

\end{abstract}

\section{Introduction}

Multilingual machine translation (MT) remains a cornerstone capability for large-scale language technologies in real-world applications. Recent large language models (LLMs), such as the GPT series \cite{openai2025gpt5} and Gemini models \cite{comanici2025gemini25pushingfrontier,deepmind2025gemini3}, have demonstrated strong translation performance across languages, motivating their adoption as unified translation systems. While closed-source models and commercial translation services continue to achieve high-quality multilingual MT, there is growing demand for open LLMs that combine competitive performance with transparency and flexible deployment.

Recent generations of open LLMs, including the Qwen3 series \cite{yang2025qwen3technicalreport} and Gemma3 models \cite{gemmateam2025gemma3technicalreport}, have substantially expanded their multilingual coverage by leveraging diverse training data. Despite these advances, their performance on multilingual MT tasks remains largely unexplored, and it is unclear which scaling strategies are most effective for adapting LLMs to such tasks. In particular, open questions remain regarding the relative contributions of model scaling versus data scaling, and how these factors interact across different adaptation stages, such as continual pretraining and instruction finetuning.

In this work, we address these gaps through a large-scale empirical study of open LLMs for multilingual MT, focusing on practical training strategies and reproducible evaluation. Our study spans 46 languages, and examines how scaling choices affect translation quality across languages and training stages. Building on the Gemma3 model family, we further develop MiLMMT-46, a series of many-to-many multilingual translation models of varying sizes, designed to provide a strong and deployable open alternative to proprietary systems. The contributions of this paper can be summarized as follows:
\begin{itemize}
\item We benchmark the latest open-source LLMs on multilingual MT across 46 widely spoken languages, covering both English-centric and Chinese-centric translation directions.
\item We systematically investigate the effects of model and data scaling across both the continual pretraining and instruction finetuning stages for multilingual MT with LLMs.
\item We publicly release MiLMMT-46, a series of many-to-many multilingual MT models supporting 46 languages, which consistently outperform other open-source alternatives and are competitive with closed-source systems such as Google Translate and Gemini 3 Pro.
\end{itemize}

\section{Related Work}

Recent multilingual LLMs typically improve MT performance through combinations of continual pretraining, supervised finetuning, and reinforcement learning \cite{DBLP:conf/iclr/Xu0SA24,alves2024toweropenmultilinguallarge,DBLP:conf/iclr/XuMKHEK25}. \citet{cui-etal-2025-multilingual} study data mixing strategies for continual pretraining and propose prioritizing parallel corpora to better align multilingual representations. Seed-X \cite{cheng2025seedxbuildingstrongmultilingual} incorporates linguist-authored chain-of-thought supervision and multi-reward reinforcement learning to capture fine-grained semantic and cultural distinctions across language directions. Tower+ \cite{rei2025towerbridginggeneralitytranslation} introduces a multi-stage post-training framework that balances translation specialization and general capabilities by integrating instruction data into pretraining and applying Group Relative Policy Optimization (GRPO) \cite{shao2024deepseekmathpushinglimitsmathematical}. Similarly, Hunyuan-MT \cite{zheng2025hunyuanmttechnicalreport,zheng2025hymt15technicalreport} applies GRPO with a multi-faceted reward design to model translation diversity and mitigate training collapse. TranslateGemma \cite{finkelstein2026translategemmatechnicalreport} shows that targeted post-training can substantially improve multilingual translation while preserving multimodal capabilities. In contrast to these works, which primarily focus on training strategies and reward design, our study systematically investigates the effects of data and model scaling on multilingual translation performance.

\section{Datasets and Baseline Settings}


\subsection{Datasets}

We conduct experiments on $46$ languages spanning a broad linguistic spectrum, with detailed language information summarized in Table~\ref{tab:langs}. We evaluate the multilingual translation performance on the FLORES+ \cite{nllb-24} and WMT24++ \cite{deutsch-etal-2025-wmt24} benchmarks. We adopt the English sentences from the WMT24++ benchmark and exclude those marked as low quality for reference-free evaluation.

\subsection{Models}\label{sec:baselines}

We evaluate the translation performance of several open-source LLMs, including the Qwen2.5 \cite{qwen2025qwen25technicalreport}, Qwen3 \cite{yang2025qwen3technicalreport}, Gemma2 \cite{gemmateam2024gemma2improvingopen} and Gemma3 \cite{gemmateam2025gemma3technicalreport} series. We also report results of several SOTA models as follows:
\begin{itemize}[leftmargin=*]
\item Google Translate: Commercial MT performance obtained via the Google Translate API\footnote{\url{https://translate.google.com/}}.
\item Gemini 2.5/3 Pro: Performance obtained via the Vertex AI API using default decoding.
\item GPT-5: Performance obtained via the OpenAI API using default decoding.
\item NLLB-54.5B: A large-scale encoder–decoder multilingual NMT model released by the No Language Left Behind project \cite{nllbteam2022languageleftbehindscaling}.
\end{itemize}

\subsection{Evaluation}

For the FLORES+ benchmark, we evaluate MT performance by spBLEU\footnote{We calculate the spBLEU scores via sacreBLEU \cite{post-2018-call} with the \texttt{flores200} tokenizer.} \cite{goyal-etal-2022-flores} and COMET\footnote{\url{https://huggingface.co/Unbabel/wmt22-comet-da}} \cite{rei-etal-2020-comet}. For the WMT24++ benchmark, we adopt two reference-free models, XCOMET\footnote{\url{https://huggingface.co/Unbabel/XCOMET-XXL}} \cite{guerreiro-etal-2024-xcomet} and COMETKiwi\footnote{\url{https://huggingface.co/Unbabel/wmt23-cometkiwi-da-xxl}} \cite{rei-etal-2023-scaling}, each of which has $10$B parameters and demonstrates high correlation with human judgments \cite{freitag-etal-2023-results}.

\section{Benchmarking Open LLMs for Multilingual Machine Translation}\label{sec:benchmark_mt}

\subsection{Tokenizer Efficiency}

Following \citet{cui-etal-2025-multilingual}, we evaluate tokenizer efficiency by comparing the tokenized length of English sentences with that of their non-English counterparts. We define the length ratio as
\begin{equation}
\text{length ratio} = \frac{\text{length}(\text{tokenizer}(y))}{\text{length}(\text{tokenizer}(x))},
\end{equation}
where $x$ and $y$ denote the English and non-English sentences, respectively. Smaller ratios indicate more efficient tokenization of non-English sentences relative to English sentences.

\begin{figure*}[h]
\centering
\includegraphics[scale=0.26]{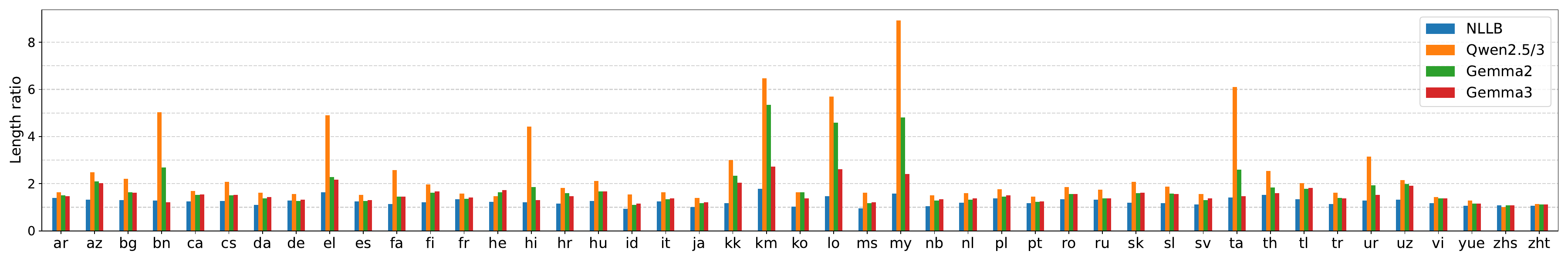} 
\caption{The tokenizer efficiency of open-source LLMs for each non-English language. The smaller the length ratio is, the more efficient the tokenizer is. The detailed results are summarized in Table \ref{tab:tokenization}.}
\label{fig:tokenization}
\end{figure*}

\begin{table*}[h]\small
\centering
\begin{tabular}{l | c | c c c c} 
\hline
\multicolumn{1}{c|}{Model} & WMT24++ & \multicolumn{4}{c}{FLORES+} \\
& \texttt{en} $\rightarrow$ \texttt{xx} & \texttt{en} $\rightarrow$ \texttt{xx} & \texttt{xx} $\rightarrow$ \texttt{en} & \texttt{zh} $\rightarrow$ \texttt{xx} & \texttt{xx} $\rightarrow$ \texttt{zh} \\
\hline
\hline
Google Translate & 83.29 / 80.46 & 42.90 / 89.86 & 47.42 / 89.42 & 30.74 / 87.46 & 36.08 / 88.24 \\
Gemini 3 Pro & 85.03 / 81.70 & 42.42 / 90.35 & 46.44 / 89.44 & 29.90 / 87.97 & 33.81 / 88.13 \\
Gemini 2.5 Pro & 84.49 / 81.46 & 41.15 / 90.07 & 46.13 / 89.38 & 29.12 / 87.76 & 33.07 / 88.01 \\
GPT-5 & 84.86 / 82.10 & 38.42 / 89.86 & 43.64 / 89.19 & 26.36 / 87.46 & 31.34 / 87.66 \\
NLLB-54.5B & - & 38.05 / 87.89 & 43.23 / 88.10 & 25.17 / 85.27 & 20.72 / 80.64 \\
\hline
Qwen2.5-0.5B & 29.90 / 13.56 & 4.86 / 47.39 & 12.90 / 66.08 & 2.66 / 46.41 & 6.74	/ 63.30 \\
Qwen2.5-1.5B & 38.90 / 27.25 & 10.77 / 58.60 & 23.89 / 77.74 & 6.54 / 56.68 & 16.35 / 75.79 \\
Qwen2.5-3B & 46.02 / 35.51 & 14.76 / 65.36 & 29.83 / 82.23 & 9.14 / 62.85 & 20.79	/ 80.22 \\
Qwen2.5-7B & 54.11 / 46.20 & 20.40 / 72.85 & 35.01 / 84.96 & 13.34 / 70.71 & 25.74 / 83.36 \\
Qwen2.5-14B & 62.31 / 56.52 & 25.32 / 79.03 & 39.52 / 87.07 & 17.01 / 76.64 & 29.28 / 85.64 \\
\hline
Qwen3-0.6B & 34.54 / 21.15 & 7.84 / 54.86 & 20.14 / 75.45 & 4.03 / 51.59 & 12.36 / 72.49 \\
Qwen3-1.7B & 47.66 / 37.71 & 15.55 / 68.96 & 30.32 / 83.52 & 9.06 / 65.85 & 20.66 / 81.22 \\
Qwen3-4B & 59.18 / 53.06 & 22.88 / 78.55 & 36.72 / 86.57 & 14.61 / 75.82 & 26.87 / 84.92 \\
Qwen3-8B & 66.28 / 61.71 & 27.93 / 83.34 & 39.59 / 87.60 & 17.88 / 80.43 & 29.30 / 86.10 \\
Qwen3-14B & 70.56 / 67.10 & 31.19 / 85.52 & 42.05 / 88.21 & 20.95 / 83.14 & 30.77 / 86.66 \\
\hline
Gemma2-2B & 57.81 / 49.92 & 22.36 / 76.72 & 34.78 / 84.87 & 12.07 / 72.28 & 19.38	/ 80.42 \\
Gemma2-9B & 72.38 / 68.76 & 33.70 / 86.23 & 43.08	/ 88.26 & 21.02	/ 82.60 & 28.79	/ 85.89 \\
\hline
Gemma3-270M & 31.65 / 14.15 & 4.68 / 53.47 & 9.02 / 66.35 & 1.78 / 49.67 & 2.96 / 56.77 \\
Gemma3-1B & 53.99 / 45.67 & 19.85 / 76.87 & 30.74 / 84.17 & 10.46 / 72.77 & 15.68 / 78.67 \\
Gemma3-4B & 71.43 / 67.39 & 31.88	/ 86.21 & 40.01 / 87.61 & 19.88 / 83.05 & 26.35 / 84.97 \\
Gemma3-12B & 78.02 / 74.62 & 38.25 / 88.66 & 44.34 / 88.73 & 25.85 / 86.03 & 31.13 / 86.84 \\
\end{tabular}
\caption{Performance of different models on WMT24++ (XCOMET / COMETKiwi) and FLORES+ (spBLEU / COMET) benchmarks. The detailed results are summarized in Tables \ref{tab:flores_plus_qwen_en}, \ref{tab:flores_plus_qwen_zh}, \ref{tab:wmt24_qwen_en}, \ref{tab:flores_plus_gemma_en}, \ref{tab:flores_plus_gemma_zh}, and \ref{tab:wmt24_gemma_en}.}
\label{tab:main_results_in-context}
\end{table*}

We conduct experiments on the FLORES+ devtest, which contains $1012$ sentences per language. We include the NLLB-54.5B tokenizer as a strong baseline. Figure \ref{fig:tokenization} shows the average length ratio for each language, and Table~\ref{tab:tokenization} additionally reports results for the Seed-X and HY-MT1.5 tokenizers. Overall, NLLB achieves consistently low length ratios across languages, while Gemma3 exhibits the most balanced tokenizer among open LLMs.

\subsection{In-context Multilingual Translation Performance with Open LLMs}

We evaluate multilingual translation of different LLMs using in-context learning on the FLORES+ benchmark. For each model, we report performance across $46$ languages with eight randomly selected translation pairs\footnote{We apply the same eight randomly selected translation pairs as exemplars for each direction during evaluation.} from the FLORES+ development dataset as the in-context exemplars. Following \citet{cui-etal-2025-multilingual}, we adopt the in-context template ``<X>=<Y>'', where <X> and <Y> denote the source and target sentences of the select parallel sentence pairs. All experiments are conducted based on OpenICL\footnote{\url{https://github.com/Shark-NLP/OpenICL}} \cite{wu-etal-2023-openicl}.

Table~\ref{tab:main_results_in-context} reports the average multilingual performance. Both closed-source and open-source LLMs show strong translation capabilities:
\begin{itemize}[leftmargin=*]
\item Recent closed-source LLMs, such as Gemini 3 Pro and GPT-5, achieve remarkable multilingual translation performance, surpassing Google Translate in most translation directions.
\item Open LLMs demonstrate steady progress in multilingual translation quality, improving from Qwen2.5 to Qwen3 and from Gemma2 to Gemma3. Gemma3 models perform best among open LLMs and even surpass strong supervised NMT systems such as NLLB-54.5B.
\end{itemize}

\section{Model and Data Scaling for Multilingual MT with Open LLMs}\label{sec:gemmax}

We investigate the effects of model and data scaling across both the continual pretraining and instruction finetuning stages for multilingual machine translation with large language models. Specifically, we continue pretraining the Gemma3 family on multilingual corpora and subsequently apply instruction finetuning using a small but high-quality parallel dataset. During instruction finetuning, we adopt the following translation prompt: Translate this from [\textit{source language}] to [\textit{target language}]:\textbackslash n[\textit{source language}]: <\textit{source sentence}>\textbackslash n[\textit{target language}]:<\textit{target sentence}>.

\begin{figure*}[h]
\centering
\includegraphics[scale=0.39]{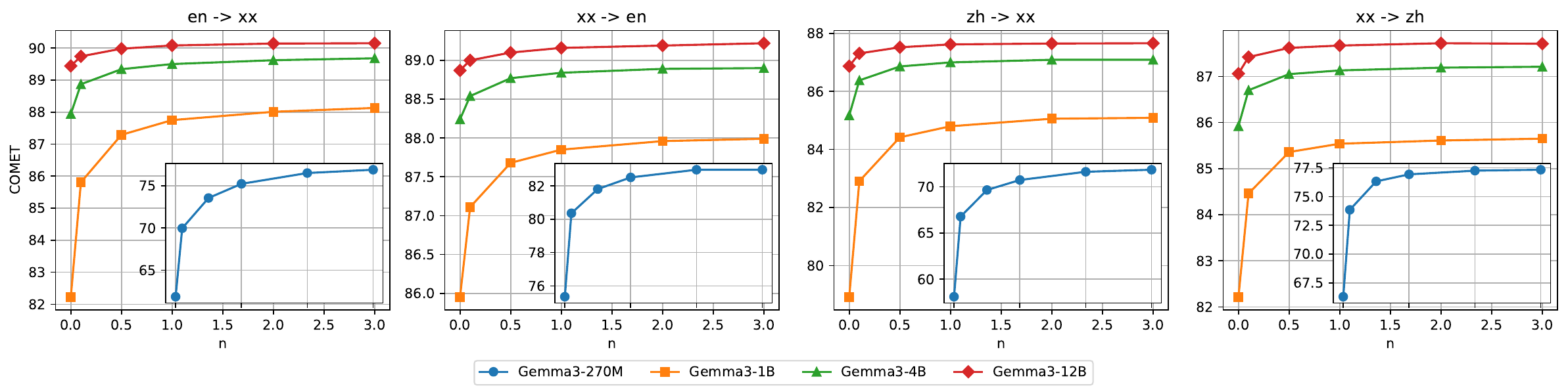} 
\caption{The translation performance (COMET) of different models trained with different $n$ during continual pretraining stage. The translation performance in BLEU scores is illustrated in Figure \ref{fig:scaling_cpt_bleu}.}
\label{fig:scaling_cpt_comet}
\end{figure*}

\subsection{Pretraining Data}

\paragraph{Monolingual Data} We collect monolingual data from DCAD-2000 \cite{shen2025dcad2000multilingualdataset2000}, which is a large-scale, high-quality multilingual dataset covering $2282$ languages, constructed via an anomaly-detection-based cleaning framework and validated across multiple multilingual benchmarks.

\paragraph{Parallel Data}
We collect all Chinese-centric and English-centric parallel datasets from the OPUS collection\footnote{\url{http://www.opus.nlpl.eu}} \cite{tiedemann-2012-parallel} released up to August 2025. All available corpora are downloaded and concatenated without manual curation or explicit domain balancing. We then apply a data-cleaning pipeline similar to that of \citet{cui-etal-2025-multilingual}, including heuristic filtering, language identification, and semantic similarity filtering. After cleaning, the resulting dataset contains approximately $4.9$ billion simplified Chinese–centric and English–centric sentence pairs spanning $46$ languages. The data distribution is shown in Figure~\ref{fig:parallel_dataset}.

\subsection{Supervised Finetuning Data}

We construct our instruction finetuning dataset from a diverse set of sources, including the FLORES+ development set, the NTREX-128 development set \cite{federmann-etal-2022-ntrex}, the TowerBlock dataset\footnote{\url{https://huggingface.co/datasets/Unbabel/TowerBlocks-v0.2}}, the BOUQuET dataset \cite{andrews-etal-2025-bouquet}, the OLDI Seed dataset\footnote{\url{https://huggingface.co/datasets/openlanguagedata/oldi_seed}}, as well as test sets from WMT15 to WMT23 \cite{bojar-etal-2015-findings,bojar-etal-2016-findings,bojar-etal-2017-findings,bojar-etal-2018-findings,barrault-etal-2019-findings,barrault-etal-2020-findings,akhbardeh-etal-2021-findings,kocmi-etal-2022-findings,kocmi-etal-2023-findings}. For each source sentence, we generate multiple candidate translations using closed-source large language models, including Gemini 3.0 Pro and GPT-5, and select the best candidate based on reference-free quality metrics, namely XCOMET and COMETKiwi. To further ensure data quality, we filter out samples with scores below a predefined threshold. The resulting finetuning dataset contains approximately $264$K sentence pairs. Detailed statistics are summarized in Table~\ref{tab:sft_data}. Overall, the dataset covers $192$ translation directions, with English-centric directions accounting for $94.5\%$ of the data, while simplified Chinese-centric directions constitute approximately $7.4\%$.



\subsection{Exploring Model and Data Scaling for Multilingual Translation with LLMs}

We train all models using the LlamaFactory framework \cite{zheng-etal-2024-llamafactory} for one epoch, with $32$ NVIDIA H100 GPUs for continual pretraining and $8$ H100 GPUs for instruction finetuning. Training configurations are summarized in Tables~\ref{tab:pretrain_setup} and \ref{tab:sft_setup}. Translations are generated using greedy decoding.

We adopt the Parallel-First Monolingual-Second (PFMS) data mixing strategy proposed by \citet{cui-etal-2025-multilingual}, which prioritizes parallel data over monolingual data when constructing the pretraining corpus. For each language, we target $n$ billion tokens, using parallel data as extensively as possible and supplementing it with monolingual data when necessary. To preserve long-context modeling capability, we additionally include $0.1n$ billion tokens of monolingual data for each language. We vary $n$ across $0.1$, $0.5$, $1$, $2$, and $3$. Dataset statistics are summarized in Tables \ref{tab:cpt_data_0.1b}, \ref{tab:cpt_data_0.5b}, \ref{tab:cpt_data_1b}, \ref{tab:cpt_data_2b} and \ref{tab:cpt_data_3b}.

We continually pretrain Gemma3 models under these five token budgets and subsequently apply instruction finetuning on the full high-quality dataset. Figure \ref{fig:scaling_cpt_comet} shows the effect of continual pretraining scale on multilingual translation performance across model sizes and translation directions. Several consistent trends are observed:
\begin{itemize}[leftmargin=*]
\item Increasing the continual pretraining data scale yields stable performance improvements across all model sizes and language directions.
\item Larger models consistently achieve higher absolute performance. However, performance gains, particularly in terms of COMET, exhibit diminishing returns as the data scale increases, especially for larger models.
\end{itemize}

\begin{figure*}[h]
\centering
\includegraphics[scale=0.39]{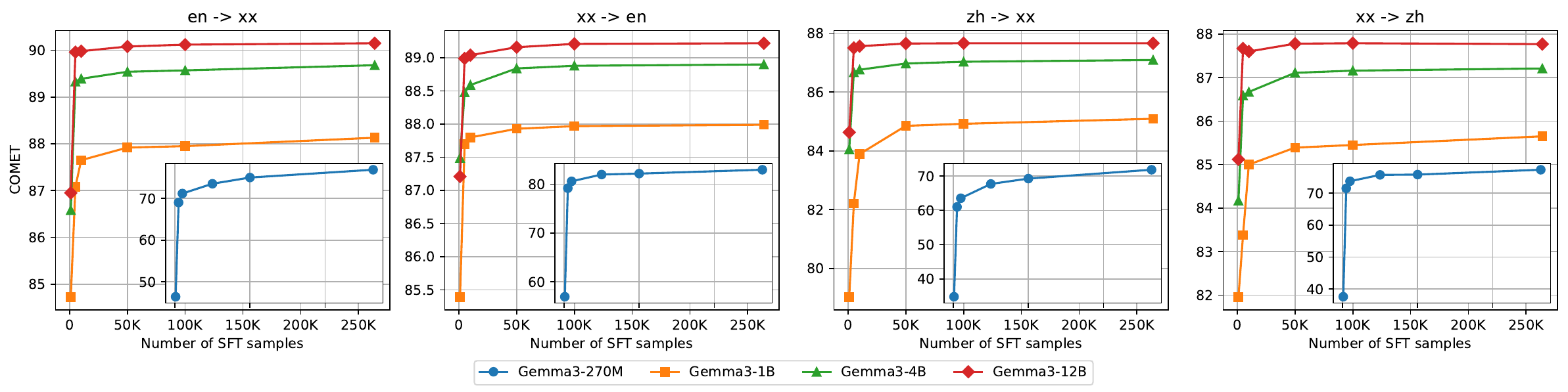} 
\caption{The translation performance (COMET) of different models trained with varying numbers of sentence pairs during the instruction finetuning stage. Translation performance measured by BLEU is shown in Figure~\ref{fig:scaling_sft_bleu}.}
\label{fig:scaling_sft_comet}
\end{figure*}

We then randomly sample $1$K, $5$K, $10$K, $50$K, and $100$K sentence pairs from the high-quality instruction finetuning dataset and finetune models that have been continually pretrained with $n=3$. Figure \ref{fig:scaling_sft_comet} shows the impact of instruction finetuning data scale on multilingual translation performance across model sizes and translation directions. Several consistent trends are observed:
\begin{itemize}[leftmargin=*]
\item Instruction finetuning consistently improves translation performance across all model sizes and translation directions, with the most pronounced gains observed when increasing the data scale from small to moderate sizes.
\item Larger models are more data-efficient during instruction finetuning, achieving strong many-to-many multilingual translation with fewer parallel sentence pairs. For the largest models, approximately $100$K high-quality pairs suffice for robust performance across all $46$ languages, with diminishing returns beyond this scale. Notably, in the $100$K setting, only $154$ sentence pairs cover \texttt{zh} $\leftrightarrow$ \texttt{xx} (excluding English) translation directions, indicating strong zero-shot translation capability.
\end{itemize}

\subsection{Main Result}

\begin{table*}[h!]\small
\centering
\begin{tabular}{l | c | c c c c} 
\hline
\multicolumn{1}{c|}{Model} & \multicolumn{1}{c|}{WMT24++} & \multicolumn{4}{c}{FLORES+} \\
& \texttt{en} $\rightarrow$ \texttt{xx} & \texttt{en} $\rightarrow$ \texttt{xx} & \texttt{xx} $\rightarrow$ \texttt{en} & \texttt{zh} $\rightarrow$ \texttt{xx} & \texttt{xx} $\rightarrow$ \texttt{zh} \\
\hline
\hline
\textbf{$21$ languages} & & \\
Tower-Plus-2B & 82.52 / 76.30 & 40.41 / 89.41 & 42.54 / 88.77 & 24.27 / 86.62 & 29.53 / 86.97 \\
Tower-Plus-9B & 86.80 / 80.76 & 43.33 / 90.31 & 45.32 / 89.35 & 27.59 / 87.71 & \textbf{33.25} / 88.06 \\
MiLMMT-46-1B & 80.92 / 74.23 & 38.14 / 88.71 & 42.17 / 88.53 & 23.62 / 85.86 & 28.31 / 86.47 \\
MiLMMT-46-4B & 86.94 / 80.63 & 41.90 / 90.05 & 44.69 / 89.16 & 27.60 / 87.60 & 31.87 / 87.75 \\
MiLMMT-46-12B & \textbf{88.54} / \textbf{82.47} & \textbf{43.51} / \textbf{90.48} & \textbf{45.96} / \textbf{89.40} & \textbf{29.43} / \textbf{88.14} & 33.16 / \textbf{88.17} \\
\hline
\hline
\textbf{$26$ languages} & & \\
Seed-X-Instruct-7B & 85.19 / 78.30 & 44.16 / 90.42 & 44.54 / 88.66 & 28.60 / 87.53 & 32.51 / 87.75 \\
Seed-X-PPO-7B & 86.18 / 79.99 & \textbf{45.48} / 90.76 & 44.12 / 89.18 & 29.65 / 88.36 & 29.20 / \textbf{88.01} \\
MiLMMT-46-4B & 86.96 / 80.98 & 42.91 / 90.49 & 45.49 / 89.22 & 27.89 / 87.94 & 31.71 / 87.50 \\
MiLMMT-46-12B & \textbf{88.56} / \textbf{82.76} & 44.49 / \textbf{90.91} & \textbf{46.77} / \textbf{89.47} & \textbf{29.72} / \textbf{88.48} & \textbf{33.06} / 87.94 \\
\hline
\hline
\textbf{$28$ languages} & & \\
GemmaX2-28-2B & 76.89 / 73.15 & 37.05 / 87.56 & 42.18 / 88.34 & 24.07 / 84.47 & 30.60 / 86.43 \\
GemmaX2-28-9B & 79.21 / 75.55 & 39.77 / 88.36 & \textbf{45.09} / 88.96 & 27.48 / 85.71 & \textbf{33.77} / 87.40 \\
MiLMMT-46-1B & 76.86 / 73.27 & 34.17 / 87.37 & 39.85 / 87.92 & 21.82 / 84.18 & 26.42 / 85.45 \\
MiLMMT-46-4B & 83.51 / 79.61 & 38.52 / 88.86 & 43.34 / 88.81 & 26.14 / 86.04 & 30.53 / 87.02 \\
MiLMMT-46-12B & \textbf{85.24} / \textbf{81.31} & \textbf{40.18} / \textbf{89.30} & 44.87 / \textbf{89.12} & \textbf{27.91} / \textbf{86.58} & 32.03 / \textbf{87.55} \\
\hline
\hline
\textbf{$31$ languages} & & \\
HY-MT1.5-1.8B & 83.88 / 77.38 & 24.81 / 86.10 & 25.25 / 85.73 & 17.60 / 82.25 & 22.61 / 86.00 \\
Hunyuan-MT-7B & 84.51 / 80.71 & 27.84 / 87.14 & 30.20 / 86.98 & 20.83 / 84.73 & 22.98 / 86.72 \\
HY-MT1.5-7B & 84.42 / 80.85 & 28.97 / 87.45 & 31.68 / 87.34 & 20.61 / 84.79 & 23.59 / 86.85 \\
MiLMMT-46-1B & 76.12 / 73.27 & 33.60 / 87.66 & 39.00 / 87.79 & 22.21 / 84.87 & 26.76 / 85.69 \\
MiLMMT-46-4B & 82.81 / 79.60 & 37.96 / 89.12 & 42.50 / 88.69 & 26.42 / 86.63 & 30.76 / 87.21 \\
MiLMMT-46-12B & \textbf{84.61} / \textbf{81.34} & \textbf{39.64} / \textbf{89.58} & \textbf{44.05} / \textbf{89.01} & \textbf{28.12} / \textbf{87.14} & \textbf{32.23} / \textbf{87.72} \\
\hline
\hline
\textbf{$46$ languages} & & \\
Google Translate & 83.29 / 80.46 & \textbf{42.90} / 89.86 & \textbf{47.42} / 89.42 & \textbf{30.74} / 87.46 & \textbf{36.08} / \textbf{88.24} \\
Gemini 3 Pro & 85.03 / 81.70 & 42.42 / \textbf{90.35} & 46.44 / \textbf{89.44} & 29.90 / \textbf{87.97} & 33.81 / 88.13 \\
Gemini 2.5 Pro & 84.49 / 81.46 & 41.15 / 90.07 & 46.13 / 89.38 & 29.12 / 87.76 & 33.07 / 88.01 \\
GPT-5 & 84.86 / \textbf{82.10} & 38.42 / 89.86 & 43.64 / 89.19 & 26.36 / 87.46 & 31.34 / 87.66 \\
NLLB-54.5B & - & 38.05 / 87.89 & 43.23 / 88.10 & 25.17 / 85.27 & 20.72 / 80.64 \\
TranslateGemma-4B & 75.97 / 73.03 & 27.71 / 85.09 & 33.42 / 87.26 & 17.71 / 82.60 & 23.37 / 85.75 \\
TranslateGemma-12B & 84.16 / 81.63 & 31.05 / 89.08 & 35.45 / 88.09 & 21.56 / 86.62 & 26.75 / 87.17 \\
MiLMMT-46-1B & 76.14 / 73.08 & 35.07 / 88.17 & 40.61 / 88.04 & 22.12 / 85.18 & 26.94 / 85.63 \\
MiLMMT-46-4B & 83.27 / 79.93 & 39.56 / 89.70 & 43.93 / 88.91 & 26.41 / 87.11 & 30.96 / 87.19 \\
MiLMMT-46-12B & \textbf{85.09} / 81.75 & 41.24 / 90.16 & 45.45 / 89.22 & 28.27 / 87.67 & 32.44 / 87.70 \\
\hline
\end{tabular}
\caption{Translation performance on WMT24++ (XCOMET / COMETKiwi) and FLORES+ (spBLEU / COMET) benchmarks. The detailed results are summarized in Tables \ref{tab:flores_plus_baseline_en}, \ref{tab:flores_plus_baselines_zh}, \ref{tab:wmt24_baseline_en}, \ref{tab:flores_plus_ours_en}, \ref{tab:flores_plus_ours_zh}, and \ref{tab:wmt24_ours_en}.}
\label{tab:main_results}
\end{table*}

We summarize the experimental results in Table \ref{tab:main_results}. All MiLMMT models are trained using the full continual pretraining corpus and instruction finetuning datasets. In addition to the models introduced in Section \ref{sec:baselines}, we compare against several strong open-source multilingual baselines:
\begin{itemize}[leftmargin=*]
\item Tower-Plus-2B/9B \cite{rei2025towerbridginggeneralitytranslation}: Gemma2-based models for multilingual translation and general-purpose tasks across $27$ languages.
\item GemmaX2-2B/9B \cite{cui-etal-2025-multilingual}: Gemma2-based models designed for multilingual machine translation across $28$ languages.
\item Seed-X-Instruct/PPO-7B \cite{cheng2025seedxbuildingstrongmultilingual}: Mistral-based models trained with instruction finetuning and reinforcement learning for multilingual machine translation across $28$ languages.
\item Hunyuan-MT-7B and HY-MT1.5-1.8B/7B \cite{zheng2025hunyuanmttechnicalreport,zheng2025hymt15technicalreport}: Hunyuan-based models for multilingual translation across $33$ languages.
\item TranslateGemma-4B/12B \cite{finkelstein2026translategemmatechnicalreport}: Gemma3-based models for high-quality translation across $55$ languages.
\end{itemize}
We evaluate each baseline only on languages that overlap with MiLMMT. Across all evaluation settings, MiLMMT achieves strong and consistent performance. In the limited language coverage scenarios ($21$, $26$, and $28$ languages), MiLMMT-46-12B attains the best overall results on almost all metrics, outperforming similarly sized open-source models such as Tower-Plus-9B, Seed-X-7B, and GemmaX2-9B on both WMT24++ and FLORES+. As coverage increases to $31$ and $46$ languages, MiLMMT continues to scale favorably, substantially surpassing Hunyuan-MT and HY-MT1.5 baselines and remaining competitive with proprietary systems such as Google Translate, Gemini 3 Pro, and GPT-5, while clearly outperforming large open-source models such as NLLB and TranslateGemma at comparable or larger scales.

\section{Conclusion}

In this paper, we study multilingual machine translation with open large language models, analyzing the effects of model and data scaling across continual pretraining and instruction finetuning. Experiments covering $46$ languages on WMT24++ and FLORES+ benchmarks show that open LLMs can achieve strong many-to-many translation performance when properly adapted. Based on the Gemma3 family, we develop MiLMMT-46, a suite of open multilingual MT models that consistently outperform open-source baselines and remain competitive with proprietary systems across varying language coverage. Our results highlight the importance of large-scale multilingual pretraining and high-quality instruction finetuning, with larger models exhibiting improved data efficiency and cross-lingual generalization. We hope our findings and released models support scalable, transparent, and deployable multilingual translation systems. In future work, we will further explore reinforcement learning to better align translations with human preferences and improve performance on challenging translation tasks.

\section*{Limitations}

Due to limited computational resources, we restrict our study to multilingual in-context translation evaluation and investigate model and data scaling effects on open LLMs with fewer than 15 billion parameters. The translation performance and scaling behavior of larger models remain unexplored.

\bibliography{custom}

\appendix

\section{Appendix}\label{sec:appendix}

\begin{table*}[h]
\centering
\begin{tabular}{c | c | c | c | c | c}
ISO Code & Language & Script & Family & Subgrouping & Resource \\
\hline
\hline
\texttt{ar} & Arabic & Arabic & Afro-Asiatic & Semitic & High \\
\texttt{az} & Azerbaijani & Latin & Turkic & Common Turkic & Low\\
\texttt{bg} & Bulgarian & Cyrillic & Indo-European & Balto-Slavic & Mid \\
\texttt{bn} & Bengali & Bengali & Indo-European & Indo-Aryan & Mid \\
\texttt{ca} & Catalan & Latin & Indo-European & Italic & High \\
\texttt{cs} & Czech & Latin &  Indo-European & Balto-Slavic & High \\
\texttt{da} & Danish & Latin & Indo-European & Germanic & Mid \\
\hline
\end{tabular}
\caption{$46$ languages supported by our model. The resource of each language is determined according to the taxonomy classes by \citet{joshi-etal-2020-state}.}
\label{tab:langs}
\end{table*}

\begin{table*}[h]
\centering
\begin{tabular}{c | c | c | c | c | c | c}
Language & NLLB & Qwen2/2.5/3 & Gemma2 & Gemma3 & Seed-X & HY-MT1.5 \\
\hline
\hline
English & 30.28 & 27.29 & 26.72 & 26.79 & 28.87 & 27.30 \\
\hline
Arabic & 1.39 & 1.63 & 1.5 & 1.47 & 2.14 & 2.99 \\
Azerbaijani & 1.33 & 2.49 & 2.1 & 2.02 & 2.63 & 2.61 \\
Bulgarian & 1.3 & 2.2 & 1.63 & 1.62 & 1.91 & 2.60 \\
Bengali & 1.28 & 5.03 & 2.69 & 1.21 & 4.43 & 5.72 \\
Catalan & 1.25 & 1.69 & 1.52 & 1.54 & 1.62 & 1.70 \\
Czech & 1.26 & 2.07 & 1.5 & 1.53 & 1.88 & 2.08 \\
\hline
\end{tabular}
\caption{Tokenization efficiency of different models, where the value for English represents the average sentence length in English.}
\label{tab:tokenization}
\end{table*}

\begin{table*}[h]
    \centering
    \resizebox{0.9\textwidth}{!}{
    \begin{tabular}{c | c c c c c | c c c c c } 
Direction & Qwen2.5-0.5B & Qwen2.5-1.5B & Qwen2.5-3B & Qwen2.5-7B & Qwen2.5-14B & Qwen3-0.6B & Qwen3-1.7B & Qwen3-4B & Qwen3-8B & Qwen3-14B \\
    \hline
    \hline
        \texttt{en} $\rightarrow$ \texttt{ar} & 20.54 / 79.12 & 12.42 / 71.81 & 18.49 / 77.91 & 26.38 / 83.12 & 32.42 / 85.23 & 5.17 / 64.27 & 14.76 / 75.98 & 24.22 / 82.14 & 30.33 / 84.27 & 33.38 / 85.77 \\
    \texttt{ar} $\rightarrow$ \texttt{en} & 15.4 / 74.53 & 30.18 / 83.74 & 36.84 / 86.02 & 40.9 / 87.04 & 45.03 / 87.89 & 20.12 / 78.03 & 33.61 / 84.56 & 39.8 / 86.65 & 43 / 87.37 & 45.59 / 87.91 \\
    \hline
    \texttt{en} $\rightarrow$ \texttt{az} & 0.22 / 27.2 & 0.34 / 27.46 & 0.59 / 34.77 & 1.72 / 50.3 & 3.28 / 59.4 & 0.31 / 28.74 & 0.95 / 43.98 & 3.51 / 62.64 & 8.67 / 74.66 & 13.57 / 79.59 \\
    \texttt{az} $\rightarrow$ \texttt{en} & 2.21 / 53.53 & 7.15 / 67.52 & 12.41 / 76 & 19.99 / 81.55 & 23.38 / 83.82 & 4.96 / 64.62 & 13.19 / 78.11 & 22.66 / 83.62 & 25.81 / 85.31 & 28.36 / 86.35 \\
    \hline
    \texttt{en} $\rightarrow$ \texttt{bg} & 1.29 / 35.68 & 5.85 / 53.52 & 12.03 / 64.62 & 21.93 / 76.45 & 29.21 / 82.85 & 5.02 / 53.17 & 18.12 / 72.87 & 29.16 / 83.72 & 35.86 / 86.94 & 38.85 / 88.56 \\
    \texttt{bg} $\rightarrow$ \texttt{en} & 12.79 / 68.39 & 29.69 / 81.57 & 35.33 / 85 & 41.35 / 86.82 & 44.27 / 87.56 & 26.23 / 80.56 & 36.75 / 85.62 & 41.67 / 87.15 & 43.86 / 87.84 & 46.11 / 88.25 \\
    \hline
    \texttt{en} $\rightarrow$ \texttt{bn} & 0.12 / 29.59 & 0.81 / 37.77 & 2.33 / 46.53 & 7.98 / 63.65 & 16.36 / 76.05 & 0.5 / 33.27 & 4.15 / 54.43 & 12.32 / 72.25 & 19.53 / 79 & 23.99 / 82.58 \\
\end{tabular}}
\caption{English-centric evaluation results (spBLEU / COMET) of Qwen series on the FLORES+ benchmark. Note that the translation performance is based on the $8$-shot in-context learning strategy.}\label{tab:flores_plus_qwen_en}
\end{table*}

\begin{table*}[h]
    \centering
    \resizebox{0.9\textwidth}{!}{
    \begin{tabular}{c | c c c c c | c c c c c } 
Direction & Qwen2.5-0.5B & Qwen2.5-1.5B & Qwen2.5-3B & Qwen2.5-7B & Qwen2.5-14B & Qwen3-0.6B & Qwen3-1.7B & Qwen3-4B & Qwen3-8B & Qwen3-14B \\
\hline
\hline
    \texttt{zhs} $\rightarrow$ \texttt{ar} & 2.33 / 55.27 & 7.15 / 68.61 & 11.86 / 74.82 & 17.39 / 80.22 & 21.84 / 82.07 & 2.04 / 59.01 & 8.1 / 72.69 & 15.33 / 78.9 & 19.74 / 81.62 & 22.47 / 82.73 \\
    \texttt{ar} $\rightarrow$ \texttt{zhs} & 6.66 / 69.77 & 19.14 / 80.62 & 23.53 / 83.02 & 27.76 / 84.81 & 30.32 / 85.52 & 10.51 / 73.69 & 21.5 / 81.24 & 27.15 / 84.39 & 29.05 / 85.23 & 31.09 / 85.86 \\
    \hline
    \texttt{zhs} $\rightarrow$ \texttt{az} & 0.13 / 26.54 & 0.21 / 25.75 & 0.41 / 30.16 & 1.12 / 46.01 & 2.27 / 54.69 & 0.2 / 26.62 & 0.61 / 39.66 & 2.19 / 57.18 & 5.75 / 69.82 & 8.72 / 76.38 \\
    \texttt{az} $\rightarrow$ \texttt{zhs} & 1.71 / 51.01 & 6.88 / 65.85 & 13.06 / 74.68 & 18.72 / 79.9 & 22.1 / 83.01 & 4.33 / 62.78 & 14.02 / 76.99 & 20.24 / 82 & 22.79 / 83.77 & 24.93 / 84.84 \\
    \hline
    \texttt{zhs} $\rightarrow$ \texttt{bg} & 0.63 / 35.67 & 2.45 / 50.2 & 5.89 / 62.44 & 13.75 / 75.72 & 17.65 / 80.8 & 1.75 / 48.08 & 7.91 / 69.15 & 17.13 / 81.07 & 21.33 / 84.65 & 24.68 / 86.45 \\
    \texttt{bg} $\rightarrow$ \texttt{zhs} & 6.07 / 64.71 & 18.69 / 78.73 & 23.52 / 82.4 & 27.96 / 84.82 & 30.9 / 85.83 & 14.29 / 75.11 & 23.39 / 82.27 & 28.87 / 85.43 & 30.76 / 85.99 & 32.27 / 86.46 \\
    \hline
    \texttt{zhs} $\rightarrow$ \texttt{bn} & 0.09 / 31.05 & 0.6 / 34.78 & 1.38 / 41.55 & 4.98 / 59.83 & 10.24 / 70.69 & 0.25 / 30.21 & 1.93 / 48.03 & 8 / 67.81 & 11.95 / 74.49 & 16.03 / 77.97 \\
\end{tabular}}
\caption{Chinese-centric evaluation results (spBLEU / COMET) of Qwen series on the FLORES+ benchmark. Note that the translation performance is based on the $8$-shot in-context learning strategy.}\label{tab:flores_plus_qwen_zh}
\end{table*}

\begin{table*}[h]\tiny
\centering
\resizebox{\textwidth}{!}{
    \begin{tabular}{c | c c c c c | c c c c c } 
Direction & Qwen2.5-0.5B & Qwen2.5-1.5B & Qwen2.5-3B & Qwen2.5-7B & Qwen2.5-14B & Qwen3-0.6B & Qwen3-1.7B & Qwen3-4B & Qwen3-8B & Qwen3-14B \\
\hline
\hline
\texttt{en} $\rightarrow$ \texttt{ar} & 29.66 / 17.37 & 47.19 / 40.04 & 58.14 / 48.40 & 67.66 / 63.21 & 74.21 / 70.35 & 35.80 / 24.55 & 53.47 / 47.68 & 65.86 / 61.07 & 71.04 / 66.53 & 74.34 / 70.76 \\
\hline
\texttt{en} $\rightarrow$ \texttt{az} & 20.94 / 2.27 & 18.64 / 4.65 & 19.00 / 7.17 & 26.02 / 18.50 & 32.08 / 28.28 & 16.90 / 3.54 & 22.63 / 13.11 & 31.47 / 27.29 & 42.07 / 42.95 & 50.03 / 53.26 \\
\hline
\texttt{en} $\rightarrow$ \texttt{bg} & 18.90 / 3.98 & 27.76 / 12.66 & 36.63 / 21.01 & 50.07 / 38.10 & 62.29 / 52.35 & 27.78 / 12.44 & 45.48 / 32.97 & 62.95 / 53.91 & 70.05 / 61.63 & 73.82 / 67.56 \\
\hline
\texttt{en} $\rightarrow$ \texttt{bn} & 12.98 / -0.20 & 15.86 / 6.77 & 24.96 / 17.25 & 38.57 / 35.94 & 53.19 / 53.87 & 15.10 / 4.42 & 26.45 / 22.14 & 46.71 / 46.70 & 56.90 / 57.75 & 64.19 / 64.50 \\
\hline
\texttt{en} $\rightarrow$ \texttt{ca} & 28.70 / 9.43 & 38.78 / 23.41 & 50.65 / 36.40 & 60.31 / 48.88 & 70.25 / 59.56 & 35.95 / 18.45 & 52.94 / 38.08 & 67.67 / 56.51 & 72.85 / 63.20 & 76.06 / 67.91 \\
\hline
\texttt{en} $\rightarrow$ \texttt{cs} & 20.09 / 4.93 & 31.85 / 18.14 & 42.67 / 31.02 & 56.95 / 46.88 & 68.68 / 60.68 & 26.57 / 11.45 & 42.96 / 29.33 & 57.98 / 47.77 & 66.12 / 58.40 & 71.43 / 63.66 \\
\hline
\texttt{en} $\rightarrow$ \texttt{da} & 32.62 / 10.01 & 44.98 / 25.68 & 57.30 / 37.49 & 66.50 / 49.90 & 75.81 / 61.95 & 41.51 / 20.25 & 59.53 / 40.80 & 71.77 / 56.71 & 77.76 / 63.73 & 81.12 / 67.91 \\
\end{tabular}}
\caption{Evaluation results (XCOMET / COMETKiwi) of Qwen series on the WMT24++ benchmark. Note that the translation performance is based on the $8$-shot in-context learning strategy.}\label{tab:wmt24_qwen_en}
\end{table*}

\begin{table*}[h]
    \centering
    \resizebox{!}{0.45\textheight}{
    \begin{tabular}{c | c c | c c c c c } 
Direction & Gemma2-2B & Gemma2-9B & Gemma3-270M & Gemma3-1B & Gemma3-4B & Gemma3-12B \\
\hline
\hline
\texttt{en} $\rightarrow$ \texttt{ar} & 20.54 / 79.12 & 35.6 / 86 & 3.64 / 58.28 & 20.68 / 80.11 & 33.38 / 85.19 & 40.23 / 87.32  \\
    \texttt{ar} $\rightarrow$ \texttt{en} & 37.81 / 85.77 & 46.34 / 88.03 & 7.83 / 66.67 & 32.83 / 84.12 & 42.83 / 87.12 & 47.79 / 88.21  \\
    \hline
    \texttt{en} $\rightarrow$ \texttt{az} & 2.24 / 57.69 & 16.62 / 83.56 & 0.28 / 34.75 & 2.25 / 58.5 & 12.35 / 80.64 & 21.07 / 86.82  \\
    \texttt{az} $\rightarrow$ \texttt{en} & 19.07 / 81.31 & 29.73 / 86.84 & 2.1 / 53.99 & 15.71 / 78.77 & 26.04 / 85.33 & 30.61 / 87.16  \\
    \hline
    \texttt{en} $\rightarrow$ \texttt{bg} & 31 / 84.83 & 44.18 / 90.34 & 4.11 / 54.4 & 28.98 / 82.75 & 43.14 / 89.82 & 48.41 / 91.13  \\
    \texttt{bg} $\rightarrow$ \texttt{en} & 41.84 / 87.01 & 47.24 / 88.53 & 10.15 / 68.89 & 36.33 / 85.68 & 45.16 / 88.16 & 48.28 / 88.74  \\
    \hline
    \texttt{en} $\rightarrow$ \texttt{bn} & 6.13 / 66.24 & 23.23 / 82.68 & 1.39 / 49.7 & 11.47 / 76.33 & 26.42 / 84.38 & 31.94 / 86.05  \\
\end{tabular}}
\caption{English-centric evaluation results (spBLEU / COMET) of Gemma series on the FLORES+ benchmark. Note that the translation performance is based on the $8$-shot in-context learning strategy.}\label{tab:flores_plus_gemma_en}
\end{table*}

\begin{table*}[h]
    \centering
    \resizebox{!}{0.45\textheight}{
    \begin{tabular}{c | c c | c c c c } 
Direction & Gemma2-2B & Gemma2-9B & Gemma3-270M & Gemma3-1B & Gemma3-4B & Gemma3-12B  \\
\hline
\hline
\texttt{zhs} $\rightarrow$ \texttt{ar} & 7.47 / 72.56 & 22.21 / 82.23 & 0.18 / 47.11 & 7.82 / 71.63 & 20.19 / 81.59 & 26.03 / 83.97  \\
    \texttt{ar} $\rightarrow$ \texttt{zhs} & 16.07 / 78.65 & 29.14 / 85.07 & 1.25 / 53.35 & 13.24 / 76.82 & 25.24 / 83.52 & 30.68 / 85.62  \\
    \hline
    \texttt{zhs} $\rightarrow$ \texttt{az} & 0.89 / 48.69 & 9.11 / 77.19 & 0.14 / 33.87 & 1.31 / 53.17 & 8.28 / 76.7 & 17 / 83.69  \\
    \texttt{az} $\rightarrow$ \texttt{zhs} & 11.54 / 76.19 & 23.11 / 84.51 & 0.65 / 46.74 & 7.85 / 73.15 & 19.57 / 82.66 & 25.5 / 85.35  \\
    \hline
    \texttt{zhs} $\rightarrow$ \texttt{bg} & 15.84 / 80.79 & 26.47 / 87.09 & 1.17 / 50.27 & 13.25 / 78.46 & 26.05 / 86.74 & 31.4 / 88.57  \\
    \texttt{bg} $\rightarrow$ \texttt{zhs} & 21.79 / 81.97 & 30.75 / 86.15 & 2.69 / 56.57 & 16.72 / 78.96 & 28.13 / 84.72 & 32.47 / 86.48  \\
    \hline
    \texttt{zhs} $\rightarrow$ \texttt{bn} & 2.26 / 55.26 & 13.8 / 77.38 & 0.19 / 41.1 & 5.78 / 69.17 & 15.03 / 79.01 & 20.92 / 81.85  \\
\end{tabular}}
\caption{Chinese-centric evaluation results (spBLEU / COMET) of Gemma series on the FLORES+ benchmark. Note that the translation performance is based on the $8$-shot in-context learning strategy.}\label{tab:flores_plus_gemma_zh}
\end{table*}

\begin{table*}[h]
\centering
\resizebox{\textwidth}{!}{
    \begin{tabular}{c | c c | c c c c } 
Direction & Gemma2-2B & Gemma2-9B & Gemma3-270M & Gemma3-1B & Gemma3-4B & Gemma3-12B  \\
\hline
\hline
\texttt{en} $\rightarrow$ \texttt{ar} & 58.40 / 52.88 & 74.32 / 70.47 & 27.97 / 14.64 & 58.21 / 51.03 & 73.40 / 68.75 & 78.54 / 74.96 \\
\hline
\texttt{en} $\rightarrow$ \texttt{az} & 25.81 / 20.89 & 56.88 / 60.86 & 22.00 / 4.39 & 26.41 / 21.32 & 52.84 / 56.10 & 68.68 / 71.56 \\
\hline
\texttt{en} $\rightarrow$ \texttt{bg} & 61.92 / 51.09 & 77.27 / 70.37 & 24.16 / 9.53 & 56.09 / 45.46 & 75.41 / 68.19 & 81.50 / 75.02 \\
\hline
\texttt{en} $\rightarrow$ \texttt{bn} & 37.64 / 34.84 & 62.90 / 63.70 & 30.62 / 11.73 & 50.22 / 45.38 & 67.11 / 65.68 & 71.74 / 71.10 \\
\hline
\texttt{en} $\rightarrow$ \texttt{ca} & 64.63 / 52.50 & 78.53 / 69.36 & 32.61 / 9.49 & 56.76 / 41.18 & 75.50 / 64.96 & 80.75 / 72.95 \\
\hline
\texttt{en} $\rightarrow$ \texttt{cs} & 59.40 / 49.00 & 76.69 / 69.51 & 25.99 / 7.84 & 48.14 / 36.51 & 72.08 / 63.08 & 79.58 / 71.96 \\
\hline
\texttt{en} $\rightarrow$ \texttt{da} & 75.69 / 60.98 & 86.21 / 74.69 & 39.62 / 17.41 & 71.31 / 56.12 & 84.68 / 73.19 & 88.32 / 77.85 \\
\end{tabular}}
\caption{Evaluation results (XCOMET / COMETKiwi) of Gemma series on the WMT24++ benchmark. Note that the translation performance is based on the $8$-shot in-context learning strategy.}\label{tab:wmt24_gemma_en}
\end{table*}

\begin{figure*}[h]
\centering
\includegraphics[scale=0.26]{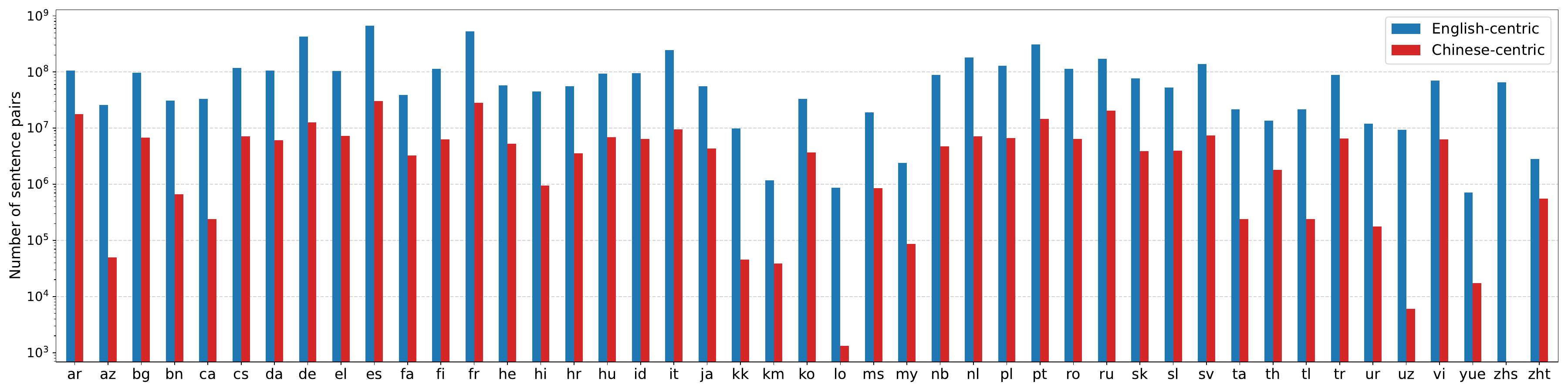} 
\caption{Number of sentence pairs for simplified Chinese-centric and English-centric parallel datasets.}
\label{fig:parallel_dataset}
\end{figure*}

\begin{table*}[t]
\footnotesize
\centering
\begin{tabular}{l c | l c | l c}
\hline
\textbf{Translation Direction} & \textbf{Count} & \textbf{Translation Direction} & \textbf{Count} & \textbf{Translation Direction} & \textbf{Count} \\
\hline
\hline
\texttt{en} $\rightarrow$ \texttt{de} & 27249 & \texttt{en} $\rightarrow$ \texttt{sk} & 1000 & \texttt{vi} $\rightarrow$ \texttt{en} & 150 \\
\texttt{de} $\rightarrow$ \texttt{en} & 23071 & \texttt{en} $\rightarrow$ \texttt{vi} & 1000 & \texttt{tl} $\rightarrow$ \texttt{en} & 150 \\
\texttt{en} $\rightarrow$ \texttt{ru} & 22112 & \texttt{en} $\rightarrow$ \texttt{tl} & 1000 & \texttt{uz} $\rightarrow$ \texttt{en} & 150 \\
\texttt{ru} $\rightarrow$ \texttt{en} & 15953 & \texttt{en} $\rightarrow$ \texttt{uz} & 1000 & \texttt{az} $\rightarrow$ \texttt{en} & 150 \\
\texttt{en} $\rightarrow$ \texttt{pt} & 10905 & \texttt{en} $\rightarrow$ \texttt{az} & 1000 & \texttt{el} $\rightarrow$ \texttt{en} & 150 \\
\texttt{en} $\rightarrow$ \texttt{zhs} & 10521 & \texttt{en} $\rightarrow$ \texttt{el} & 1000 & \texttt{my} $\rightarrow$ \texttt{en} & 150 \\
\texttt{zhs} $\rightarrow$ \texttt{en} & 8683 & \texttt{en} $\rightarrow$ \texttt{my} & 1000 & \texttt{sv} $\rightarrow$ \texttt{en} & 150 \\
\hline
\end{tabular}
\caption{Number of sentence pairs for each translation direction in the finetuning dataset.}
\label{tab:sft_data}
\end{table*}

\begin{table}[h]\small
\centering
\begin{tabular}{l|l}
\hline
Training batch size on each device & 2 \\
Number of GPUs & 32 \\
Gradient Accumulation Steps & 6 \\
Maximum Sequence Length & 4096 \\
Number of Epochs & 1 \\
Learning Rate & 2e-5\\
LR Scheduler & cosine \\
Finetuning Type & full \\
bf16 & true \\
Template & empty \\
Maximum Gradient Norm & 1.0 \\
Warmup & 0.01 \\
Weight Decay & 0.01 \\
Optimizer & AdamW \\
DeepSpeed & ZeRO2 \\
\hline
\end{tabular}
\caption{Hyperparameter settings for the pretraining experiments.}
\label{tab:pretrain_setup}
\end{table}

\begin{table}[h]\small
\centering
\begin{tabular}{l|l}
\hline
Training batch size on each device & 2 \\
Number of GPUs & 8 \\
Gradient Accumulation Steps & 16 \\
Maximum Sequence Length & 4096 \\
Number of Epochs & 1 \\
Learning Rate & 2e-5\\
LR Scheduler & inverse sqrt \\
Finetuning Type & full \\
bf16 & true \\
Template & empty \\
Maximum Gradient Norm & 1.0 \\
Warmup & 0.01 \\
Weight Decay & 0.01 \\
Optimizer & AdamW \\
DeepSpeed & ZeRO2 \\
\hline
\end{tabular}
\caption{Hyperparameter settings for the finetuning experiments.}
\label{tab:sft_setup}
\end{table}

\begin{table*}[h]\small
\centering
\begin{tabular}{c | c | c | c}
Language & \# Monolingual tokens & \# Parallel tokens (English-centric) & \# Parallel tokens (Simplified Chinese-centric) \\
\hline
\hline
\texttt{ar}  & 9999972  & 50000139  & 50000038 \\
\hline
\texttt{az}  & 10002260 & 97555558  & 2444557  \\
\hline
\texttt{bg}  & 9999929  & 50000083  & 50000021 \\
\hline
\texttt{bn}  & 9999919  & 82533473  & 17466639 \\
\hline
\texttt{ca}  & 10000145 & 91049510  & 8950662  \\
\hline
\texttt{cs}  & 10000115 & 50000179  & 50000001 \\
\hline
\texttt{da}  & 10000205 & 50000142  & 50000017 \\
\hline
\end{tabular}
\caption{Statistics for all datasets utilized during the continual pretraining stage ($n=0.1$).}
\label{tab:cpt_data_0.1b}
\end{table*}

\begin{table*}[h]\small
\centering
\begin{tabular}{c | c | c | c}
Language & \# Monolingual tokens & \# Parallel tokens (English-centric) & \# Parallel tokens (Simplified Chinese-centric) \\
\hline
\hline
\texttt{ar}  & 50000083  & 250000072 & 250000042 \\
\hline
\texttt{az}  & 50001369  & 497555460 & 2444650   \\
\hline
\texttt{bg}  & 50001866  & 252557663 & 247442490 \\
\hline
\texttt{bn}  & 50000793  & 482533496 & 17466639  \\
\hline
\texttt{ca}  & 50002382  & 491049638 & 8950468   \\
\hline
\texttt{cs}  & 50003256  & 257238346 & 242761784 \\
\hline
\texttt{da}  & 50000456  & 296376164 & 203623938 \\
\hline
\end{tabular}
\caption{Statistics for all datasets utilized during the continual pretraining stage ($n=0.5$).}
\label{tab:cpt_data_0.5b}
\end{table*}

\begin{table*}[h]\small
\centering
\begin{tabular}{c | c | c | c}
Language & \# Monolingual tokens & \# Parallel tokens (English-centric) & \# Parallel tokens (Simplified Chinese-centric) \\
\hline
\hline
\texttt{ar}  & 100000919  & 500000133  & 500000022  \\
\hline
\texttt{az}  & 100000182  & 997555642  & 2444540    \\
\hline
\texttt{bg}  & 99999858   & 752557694  & 247442454  \\
\hline
\texttt{bn}  & 130562827  & 951970867  & 17466639   \\
\hline
\texttt{ca}  & 99999879   & 991049637  & 8950564    \\
\hline
\texttt{cs}  & 99999941   & 757238324  & 242761784  \\
\hline
\texttt{da}  & 100000099  & 796375778  & 203624349  \\
\hline
\end{tabular}
\caption{Statistics for all datasets utilized during the continual pretraining stage ($n=1$).}
\label{tab:cpt_data_1b}
\end{table*}

\begin{table*}[h]\small
\centering
\begin{tabular}{c | c | c | c}
Language & \# Monolingual tokens & \# Parallel tokens (English-centric) & \# Parallel tokens (Simplified Chinese-centric) \\
\hline
\hline
\texttt{ar}  & 199999873  & 1000000114 & 1000000018 \\
\hline
\texttt{az}  & 964223959  & 1233331448 & 2444655    \\
\hline
\texttt{bg}  & 200001144  & 1752557811 & 247442302  \\
\hline
\texttt{bn}  & 1230563279 & 951970867  & 17466639   \\
\hline
\texttt{ca}  & 200000314  & 1991049698 & 8950411    \\
\hline
\texttt{cs}  & 200000243  & 1757238358 & 242761784  \\
\hline
\texttt{da}  & 200001653  & 1796376558 & 203623602  \\
\hline
\end{tabular}
\caption{Statistics for all datasets utilized during the continual pretraining stage ($n=2$).}
\label{tab:cpt_data_2b}
\end{table*}

\begin{table*}[h]\small
\centering
\begin{tabular}{c | c | c | c}
Language & \# Monolingual tokens & \# Parallel tokens (English-centric) & \# Parallel tokens (Simplified Chinese-centric) \\
\hline
\hline
\texttt{ar}  & 300000275  & 1647748942 & 1352251189 \\
\hline
\texttt{az}  & 2064228161 & 1233330310 & 2444504    \\
\hline
\texttt{bg}  & 300000904  & 2752556834 & 247443272  \\
\hline
\texttt{bn}  & 2330562536 & 951970867  & 17466639   \\
\hline
\texttt{ca}  & 1271983926 & 2019066142 & 8950542    \\
\hline
\texttt{cs}  & 300001069  & 2757238394 & 242761784  \\
\hline
\texttt{da}  & 300012930  & 2796376022 & 203624181  \\
\hline
\end{tabular}
\caption{Statistics for all datasets utilized during the continual pretraining stage ($n=3$).}
\label{tab:cpt_data_3b}
\end{table*}

\begin{figure*}[h]
\centering
\includegraphics[scale=0.39]{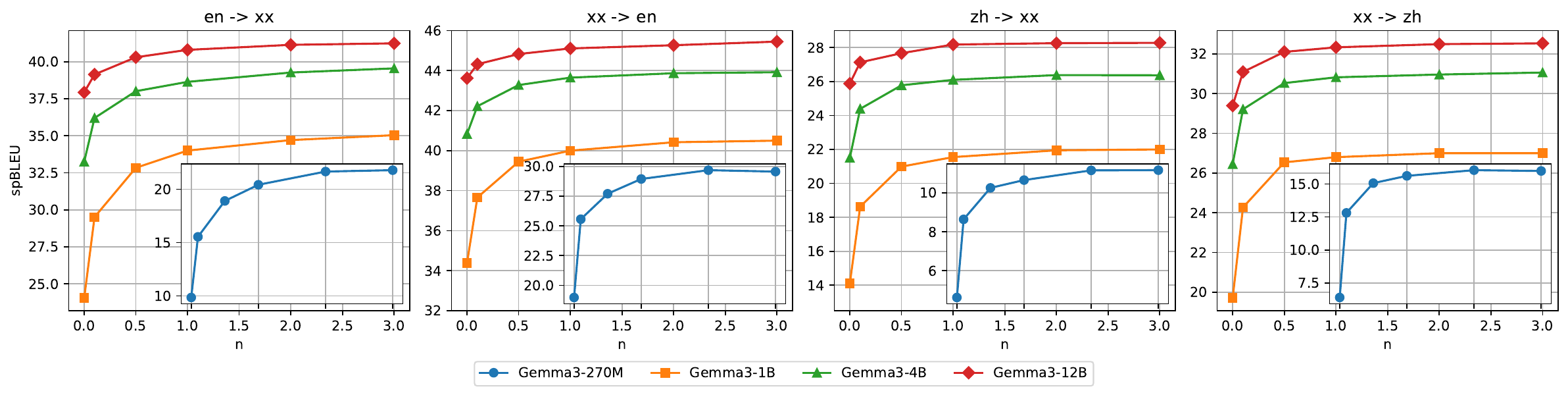} 
\caption{The translation performance (spBLEU) of different models trained with different $n$ during continual pretraining stage.}
\label{fig:scaling_cpt_bleu}
\end{figure*}

\begin{figure*}[h]
\centering
\includegraphics[scale=0.39]{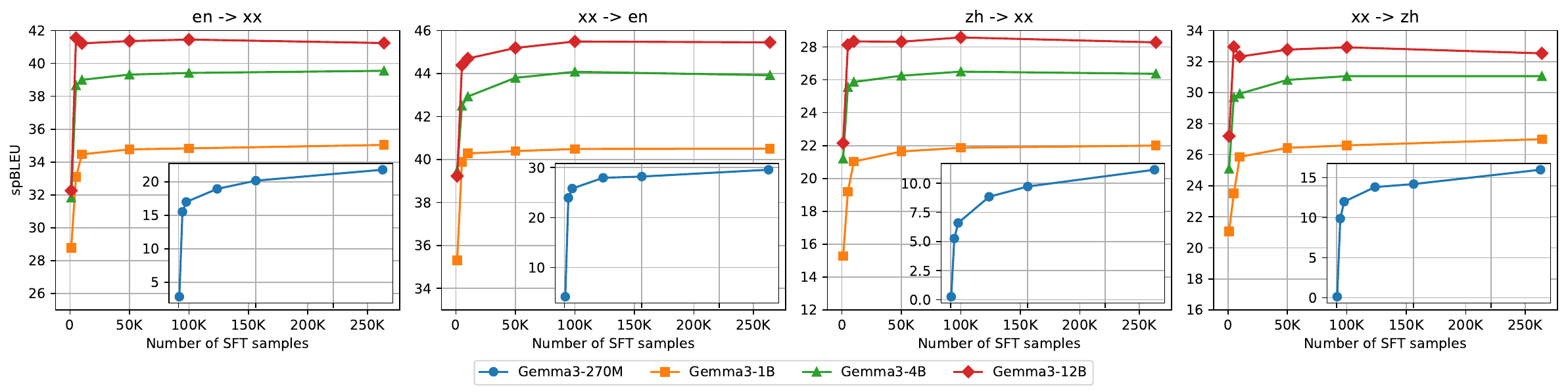} 
\caption{The translation performance (spBLEU) of different models trained with varying numbers of sentence pairs during instruction finetuning stage.}
\label{fig:scaling_sft_bleu}
\end{figure*}

\begin{table*}[h]
\centering
\resizebox{\textwidth}{!}{
\begin{tabular}{c | c c | c c | c c c | c c | c c } 
Direction & TranslateGemma-4B & TranslateGemma-12B & GemmaX2-2B & GemmaX2-9B & Hunyuan-MT-7B & HY-MT1.5-1.8B & HY-MT1.5-7B & Seed-X-Instruct-7B & Seed-X-PPO-7B & Tower-PLUS-2B & Tower-PLUS-9B \\
\hline
\hline
    \texttt{en} $\rightarrow$ \texttt{ar}  & 31.42 / 87.2 & 33.73 / 88.24 & 38.47 / 86.45 & 43 / 87.77 & 32.27 / 87.99 & 27.96 / 87.2 & 32.68 / 88.37 & 41.78 / 88.07 & 42.62 / 88.35 & -      & - \\
    \texttt{ar} $\rightarrow$ \texttt{en}  & 34.36 / 86.69 & 36.37 / 87.36 & 44.99 / 87.64 & 49.44 / 88.61 & 32.52 / 86.9 & 26.51 / 85.37 & 33.42 / 87.17 & 47.05 / 88.22 & 43.72 / 88.27 & -      & - \\
    \hline
    \texttt{en} $\rightarrow$ \texttt{az}  & 6.82 / 71.5 & 18.27 / 87.22 & -      & -      & -      & -      & -      & -      & -      & -      & - \\
    \texttt{az} $\rightarrow$ \texttt{en}  & 20.84 / 85.07 & 23.68 / 86.63 & -      & -      & -      & -      & -      & -      & -      & -      & - \\
    \hline
    \texttt{en} $\rightarrow$ \texttt{bg}  & 37.72 / 90.43 & 39.33 / 91.58 & -      & -      & -      & -      & -      & -      & -      & -      & - \\
    \texttt{bg} $\rightarrow$ \texttt{en}  & 37.76 / 87.49 & 39.51 / 88.07 & -      & -      & -      & -      & -      & -      & -      & -      & - \\
    \hline
    \texttt{en} $\rightarrow$ \texttt{bn}  & 24.35 / 86.85 & 24.81 / 87.69 & 33.14 / 86.22 & 35.03 / 86.65 & 24.89 / 87.54 & 22.67 / 86.82 & 25.77 / 88.05 & -      & -      & -      & - \\
\end{tabular}}
\caption{English-centric evaluation results (spBLEU / COMET) of baseline models on the FLORES+ benchmark.}\label{tab:flores_plus_baseline_en}
\end{table*}

\begin{table*}[h]
\centering
\resizebox{\textwidth}{!}{
    \begin{tabular}{c | c c | c c | c c c | c c | c c } 
Direction & TranslateGemma-4B & TranslateGemma-12B & GemmaX2-2B & GemmaX2-9B & Hunyuan-MT-7B & HY-MT1.5-1.8B & HY-MT1.5-7B & Seed-X-Instruct-7B & Seed-X-PPO-7B & Tower-PLUS-2B & Tower-PLUS-9B \\
\hline
\hline
    \texttt{zhs} $\rightarrow$ \texttt{ar} & 20.57 / 84.02 & 22.4 / 84.95 & 22.76 / 82.54 & 27.47 / 84.38 & 23.1 / 85.12 & 20.05 / 84.06 & 22.37 / 85.13 & 26.36 / 84.51 & 26.77 / 85.34 & -      & - \\
    \texttt{ar} $\rightarrow$ \texttt{zhs} & 23.42 / 84.56 & 26.12 / 86.06 & 30.24 / 85.18 & 34.31 / 86.45 & 22.93 / 86.08 & 22.4 / 85.26 & 24.11 / 86.14 & 31.85 / 86.44 & 27.59 / 86.69 & -      & - \\
    \hline
    \texttt{zhs} $\rightarrow$ \texttt{az} & 4.75 / 68.53 & 15.04 / 84.51 & -      & -      & -      & -      & -      & -      & -      & -      & - \\
    \texttt{az} $\rightarrow$ \texttt{zhs} & 18.51 / 83.39 & 21.47 / 85.39 & -      & -      & -      & -      & -      & -      & -      & -      & - \\
    \hline
    \texttt{zhs} $\rightarrow$ \texttt{bg} & 23.72 / 88.01 & 25.9 / 89.36 & -      & -      & -      & -      & -      & -      & -      & -      & - \\
    \texttt{bg} $\rightarrow$ \texttt{zhs} & 26.16 / 85.84 & 28.24 / 86.88 & -      & -      & -      & -      & -      & -      & -      & -      & - \\
    \hline
    \texttt{zhs} $\rightarrow$ \texttt{bn} & 16.84 / 82.43 & 18.08 / 83.51 & 19.77 / 81.36 & 24.53 / 82.77 & 18.16 / 83.43 & 16.49 / 82.47 & 17.86 / 83.71 & -      & -      & -      & - \\
\end{tabular}}
\caption{Chinese-centric evaluation results (spBLEU / COMET) of baseline models on the FLORES+ benchmark.}\label{tab:flores_plus_baselines_zh}
\end{table*}

\begin{table*}[h]
\setlength{\tabcolsep}{2pt}
\centering
\resizebox{\textwidth}{!}{
    \begin{tabular}{c | c c | c c | c c c | c c | c c } 
Direction & TranslateGemma-4B & TranslateGemma-12B & GemmaX2-2B & GemmaX2-9B & Hunyuan-MT-7B & HY-MT1.5-1.8B & HY-MT1.5-7B & Seed-X-Instruct-7B & Seed-X-PPO-7B & Tower-PLUS-2B & Tower-PLUS-9B \\
\hline
\hline
\texttt{en} $\rightarrow$ \texttt{ar} & 82.93 / 78.21 & 86.54 / 81.66 & 74.69 / 70.94 & 78.24 / 74.25 & 88.87 / 82.85 & 85.12 / 78.51 & 88.98 / 82.66 & 80.33 / 75.78 & 81.80 / 77.34 & - / - & - / - \\
\hline
\texttt{en} $\rightarrow$ \texttt{az} & 39.43 / 45.81 & 71.65 / 76.53 & - / - & - / - & - / - & - / - & - / - & - / - & - / - & - / - & - / - \\
\hline
\texttt{en} $\rightarrow$ \texttt{bg} & 83.19 / 77.11 & 88.78 / 83.66 & - / - & - / - & - / - & - / - & - / - & - / - & - / - & - / - & - / - \\
\hline
\texttt{en} $\rightarrow$ \texttt{bn} & 78.79 / 75.57 & 84.04 / 79.68 & 70.63 / 71.24 & 70.08 / 71.28 & 86.50 / 80.30 & 83.05 / 75.73 & 86.77 / 80.81 & - / - & - / - & - / - & - / - \\
\hline
\texttt{en} $\rightarrow$ \texttt{ca} & 79.59 / 71.97 & 85.65 / 78.91 & - / - & - / - & - / - & - / - & - / - & - / - & - / - & - / - & - / - \\
\hline
\texttt{en} $\rightarrow$ \texttt{cs} & 80.07 / 75.12 & 86.54 / 81.74 & 77.66 / 70.26 & 81.38 / 74.33 & 88.05 / 82.53 & 82.61 / 75.88 & 87.58 / 82.28 & 83.43 / 76.69 & 84.35 / 77.67 & 76.91 / 69.83 & 83.61 / 77.82 \\
\hline
\texttt{en} $\rightarrow$ \texttt{da} & 90.04 / 81.51 & 92.74 / 85.16 & - / - & - / - & - / - & - / - & - / - & 90.64 / 80.69 & 90.67 / 81.25 & 87.69 / 77.84 & 89.94 / 80.45 \\
\end{tabular}}
\caption{Evaluation results (XCOMET / COMETKiwi) of baseline models on the WMT24++ benchmark.}\label{tab:wmt24_baseline_en}
\end{table*}

\begin{table*}[h]
\centering
\resizebox{!}{0.45\textheight}{
    \begin{tabular}{c | c c c c c | c c c } 
Direction & Google Translate & NLLB-54.5B & Gemini 2.5 Pro & Gemini 3 Pro & GPT-5 &  MiLMMT-1B & MiLMMT-4B & MiLMMT-12B \\
\hline
\hline
    \texttt{en} $\rightarrow$ \texttt{ar} & 44.9 / 88.92 & 43.03 / 87.22 & 41.55 / 88.18 & 43.6 / 88.6 & 38.68 / 88.16 & 34.99 / 86.22 & 40.31 / 88.01 & 42.61 / 88.35 \\
    \texttt{ar} $\rightarrow$ \texttt{en} & 50.99 / 89.15 & 48.26 / 88.13 & 49.55 / 89 & 49.26 / 89.06 & 46.74 / 88.8 & 42.43 / 87.24 & 46.94 / 88.34 & 48.92 / 88.78 \\
    \hline
    \texttt{en} $\rightarrow$ \texttt{az} & 26.51 / 87.47 & 24.63 / 86.58 & 27.41 / 89.34 & 28.07 / 89.52 & 25.85 / 89.29 & 22.27 / 86.69 & 25.64 / 88.76 & 26.95 / 89.37 \\
    \texttt{az} $\rightarrow$ \texttt{en} & 33.24 / 88.13 & 31.94 / 87.33 & 31.77 / 88.07 & 32.12 / 88.09 & 29.73 / 87.81 & 26.22 / 86.08 & 29.18 / 87.23 & 30.49 / 87.58 \\
    \hline
    \texttt{en} $\rightarrow$ \texttt{bg} & 53.44 / 92.1 & 49.98 / 91.02 & 51.39 / 92.24 & 52.29 / 92.51 & 47.61 / 92.15 & 45.61 / 90.49 & 50.79 / 91.94 & 51.87 / 92.22 \\
    \texttt{bg} $\rightarrow$ \texttt{en} & 51.32 / 89.27 & 47.01 / 88.28 & 49.37 / 89.09 & 50.52 / 89.25 & 47.86 / 89.03 & 45.53 / 88.1 & 48.59 / 88.82 & 49.75 / 89.12 \\
    \hline
    \texttt{en} $\rightarrow$ \texttt{bn} & 37.55 / 88.31 & 36.04 / 87.08 & 33.47 / 87.86 & 32.86 / 88.3 & 28.46 / 87.78 & 27.62 / 86.66 & 31.66 / 88.05 & 32.73 / 88.28 \\
\end{tabular}}
\caption{English-centric evaluation results (spBLEU / COMET) of baseline models and MiLMMT models on the FLORES+ benchmark.}\label{tab:flores_plus_ours_en}
\end{table*}

\begin{table*}[h]
\centering
\resizebox{!}{0.45\textheight}{
    \begin{tabular}{c | c c c c c | c c c } 
Direction & Google Translate & NLLB-54.5B & Gemini 2.5 Pro & Gemini 3 Pro & GPT-5  & MiLMMT-1B & MiLMMT-4B & MiLMMT-12B \\
\hline
\hline
\texttt{zhs} $\rightarrow$ \texttt{ar} & 31.72 / 85.66 & 27.53 / 84.08 & 29.47 / 85.38 & 30.7 / 85.69 & 26.38 / 84.72 & 21.96 / 82.8 & 26.33 / 84.6 & 28.77 / 85.41 \\
    \texttt{ar} $\rightarrow$ \texttt{zhs} & 37.23 / 87.51 & 21.25 / 80.25 & 34.46 / 87.21 & 35.09 / 87.38 & 32.71 / 86.84 & 26.16 / 84.29 & 30.85 / 86.07 & 32.55 / 86.6 \\
    \hline
    \texttt{zhs} $\rightarrow$ \texttt{az} & 21.35 / 85.2 & 19.24 / 83.93 & 22.51 / 86.87 & 22.9 / 87.04 & 20.43 / 86.76 & 16.86 / 83.63 & 19.86 / 86.01 & 21.21 / 86.75 \\
    \texttt{az} $\rightarrow$ \texttt{zhs} & 30.41 / 87.01 & 19.32 / 81.15 & 27.77 / 86.63 & 28.12 / 86.75 & 25.42 / 86.18 & 21.06 / 83.44 & 25.17 / 85.56 & 26.56 / 86.08 \\
    \hline
    \texttt{zhs} $\rightarrow$ \texttt{bg} & 36.71 / 90.09 & 30.6 / 87.97 & 34.66 / 90.17 & 35.53 / 90.34 & 30.27 / 89.7 & 26.84 / 87.62 & 31.48 / 89.39 & 33.28 / 89.95 \\
    \texttt{bg} $\rightarrow$ \texttt{zhs} & 37.56 / 87.98 & 20.07 / 79.1 & 34.37 / 87.69 & 35.17 / 87.88 & 32.55 / 87.25 & 28.37 / 85.5 & 32.38 / 86.94 & 33.9 / 87.32 \\
    \hline
    \texttt{zhs} $\rightarrow$ \texttt{bn} & 28.29 / 83.88 & 23.67 / 81.96 & 24.99 / 84.01 & 25.28 / 84.64 & 21.08 / 83.66 & 17.26 / 81.68 & 21.89 / 83.64 & 23.36 / 84.11 \\
\end{tabular}}
\caption{Chinese-centric evaluation results (spBLEU / COMET) of baseline models and MiLMMT models on the FLORES+ benchmark.}\label{tab:flores_plus_ours_zh}
\end{table*}

\begin{table*}[h]
\centering
\resizebox{\textwidth}{!}{
    \begin{tabular}{c | c c c c | c c c} 
Direction & Google Translate & Gemini 2.5 Pro & Gemini 3 Pro & GPT-5 & MiLMMT-1B & MiLMMT-4B & MiLMMT-12B \\
\hline
\hline
   \texttt{en} $\rightarrow$ \texttt{ar} & 86.44 / 81.97 & 84.85 / 81.08 & 85.10 / 81.20 & 84.36 / 81.56 & 74.32 / 71.17 & 82.17 / 78.59 & 83.68 / 80.37 \\
    \hline
    \texttt{en} $\rightarrow$ \texttt{az} & 75.04 / 78.02 & 78.83 / 81.72 & 80.01 / 82.17 & 80.36 / 82.29 & 66.15 / 69.98 & 77.81 / 79.42 & 80.36 / 81.29 \\
    \hline
    \texttt{en} $\rightarrow$ \texttt{bg} & 85.07 / 80.04 & 86.98 / 81.51 & 87.13 / 81.84 & 87.24 / 82.49 & 78.31 / 72.59 & 85.18 / 79.44 & 87.18 / 81.96 \\
    \hline
    \texttt{en} $\rightarrow$ \texttt{bn} & 76.65 / 76.64 & 81.98 / 79.63 & 84.39 / 81.09 & 82.14 / 80.01 & 73.35 / 72.87 & 81.13 / 79.21 & 82.97 / 80.67 \\
    \hline
    \texttt{en} $\rightarrow$ \texttt{ca} & 82.82 / 76.02 & 84.16 / 77.08 & 84.97 / 77.73 & 85.39 / 78.61 & 79.25 / 70.51 & 84.88 / 77.31 & 85.82 / 78.49 \\
    \hline
    \texttt{en} $\rightarrow$ \texttt{cs} & 83.42 / 77.94 & 85.82 / 80.02 & 85.95 / 80.36 & 86.08 / 81.57 & 74.07 / 66.96 & 82.84 / 77.04 & 85.99 / 80.36 \\
    \hline
    \texttt{en} $\rightarrow$ \texttt{da} & 90.57 / 81.72 & 91.88 / 83.64 & 92.15 / 84.16 & 92.06 / 84.59 & 86.69 / 76.36 & 90.86 / 81.85 & 92.14 / 83.36 \\
\end{tabular}}
\caption{Evaluation results (XCOMET / COMETKiwi) of baseline models and MiLMMT models on the WMT24++ benchmark.}\label{tab:wmt24_ours_en}
\end{table*}

\end{document}